\pdfoutput=1
\documentclass[11pt]{article}

\usepackage[preprint]{acl}

\usepackage{times}
\usepackage{latexsym}
\usepackage[T1]{fontenc}
\usepackage[utf8]{inputenc}
\usepackage{microtype}
\usepackage{inconsolata}
\usepackage{graphicx}

\usepackage{url}            
\usepackage{booktabs}       
\usepackage{amsfonts}       
\usepackage{nicefrac}       
\usepackage{xcolor}         
\usepackage{array}
\usepackage{arydshln}
\usepackage{multirow}

\usepackage{algorithm}
\usepackage{algpseudocode}
\usepackage{amsmath}
\usepackage{geometry} 
\newcommand{\thickhline}{\noalign{\hrule height 1.2pt}}
\usepackage{tabularx} 
\usepackage{fancyvrb}
\usepackage{subcaption}
\usepackage{adjustbox}
\usepackage{float}
\usepackage{longtable} 
\usepackage{enumitem}
\usepackage{makecell}
\usepackage{hyperref} 
\usepackage{stfloats}

\title{LLM-guided Plan and Retrieval: A Strategic Alignment for Interpretable User Satisfaction Estimation in Dialogue}

\author{
    \textbf{Sangyeop Kim\textsuperscript{1,2}}, 
    \textbf{Sohhyung Park\textsuperscript{1}}, 
    \textbf{Jaewon Jung\textsuperscript{1}},
    \textbf{Jinseok Kim\textsuperscript{1}},
    \textbf{Sungzoon Cho\textsuperscript{$\dagger$1}}\\
    \textsuperscript{1} Seoul National University \\
    \textsuperscript{2} Coxwave \\
    \normalsize\texttt{\{sy917kim, sohhyung, wjdwodnjs302, jsk0821\}@bdai.snu.ac.kr}\\
    \normalsize\texttt{zoon@snu.ac.kr}
}

\begin{document}
\maketitle

\renewcommand{\thefootnote}{$\dagger$}
\footnotetext{Corresponding author.}
\renewcommand{\thefootnote}{\arabic{footnote}}

\begin{abstract}
Understanding user satisfaction with conversational systems, known as User Satisfaction Estimation (USE), is essential for assessing dialogue quality and enhancing user experiences. However, existing methods for USE face challenges due to limited understanding of underlying reasons for user dissatisfaction and the high costs of annotating user intentions. To address these challenges, we propose PRAISE (Plan and Retrieval Alignment for Interpretable Satisfaction Estimation), an interpretable framework for effective user satisfaction prediction. PRAISE operates through three key modules. The Strategy Planner develops strategies, which are natural language criteria for classifying user satisfaction. The Feature Retriever then incorporates knowledge on user satisfaction from Large Language Models (LLMs) and retrieves relevance features from utterances. Finally, the Score Analyzer evaluates strategy predictions and classifies user satisfaction. Experimental results demonstrate that PRAISE achieves state-of-the-art performance on three benchmarks for the USE task. Beyond its superior performance, PRAISE offers additional benefits. It enhances interpretability by providing instance-level explanations through effective alignment of utterances with strategies. Moreover, PRAISE operates more efficiently than existing approaches by eliminating the need for LLMs during the inference phase.
\end{abstract}

\section{Introduction}
Dialogue systems play an increasingly crucial role in enabling users to interact with intelligent agents to fulfill their needs. \textbf{User Satisfaction Estimation} (USE), the process of predicting how satisfied a user is in dialogue interactions \cite{choi2019offline}, is crucial for evaluating the quality of dialogue systems and ensuring a positive user experience \cite{bodigutla2019multidomain, Cai-2020,usda, liang-etal-2021-turn, pan-etal-2022-user, siro2023understanding, asap}. Effective USE methods should not only accurately classify user satisfaction but also provide interpretable results to guide the improvement of dialogue systems. By understanding and quantifying user satisfaction, dialogue systems can be continuously improved to better meet user expectations.

The development of USE methods has evolved through three main approaches: content-based, dialogue act-based, and language model-based. Content-based methods, such as sentiment analysis \cite{song-etal-2019} and response quality assessment \cite{SCHMITT2015}, evaluate dialogue content to estimate user satisfaction. However, these methods often struggle to accurately capture user intentions and whether user goals are fulfilled.

Dialogue act-based methods incorporate dialogue acts, which represent user intentions at each turn \cite{chen2018dialogue, stolcke2000dialogue, asap, yu2019modeling}, leveraging the relationship between these acts and user goal achievement \cite{usda}. However, these methods often require complex pre-training procedures and accurate dialogue act labels, which can be challenging and time-consuming to obtain for real-world conversations.

Language model-based methods have shown a promising direction for USE. \citet{spur} employs Large Language Models (LLMs) and iterative prompting to summarize dialogues and extract rubrics of user satisfaction from natural language utterances. However, this approach does not provide utterance-level interpretability and relies heavily on advanced LLMs like GPT-4 for the entire process, which can be expensive and impractical for large-scale applications.

\begin{figure}[!htbp]
    \centering
    \includegraphics[width=1.0\linewidth]{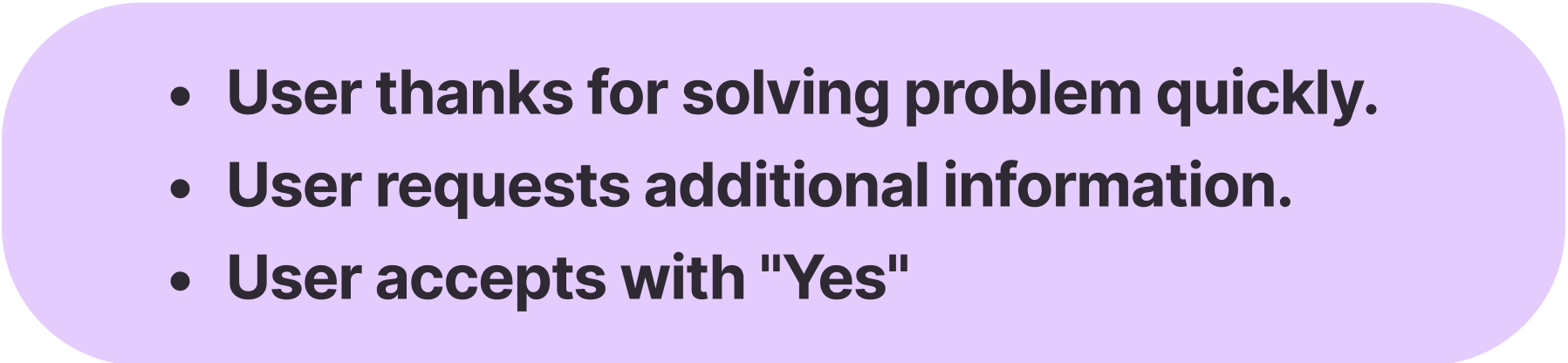}
    \caption{Examples of strategies for satisfaction}
    \label{fig:strategies}
\end{figure}

In this paper, we introduce \textbf{PRAISE} (\textbf{P}lan and \textbf{R}etrieval \textbf{A}lignment for \textbf{I}nterpretable \textbf{S}atisfaction \textbf{E}stimation), a framework that leverages LLMs to generate and refine strategies for classifying user satisfaction levels in conversational systems. These strategies serve as interpretable natural language criteria that indicate situations for identifying user satisfaction (SAT), dissatisfaction (DSAT) or neutrality (NEU) in dialogues. For instance, as shown in Figure \ref{fig:strategies}, a strategy like "User thanks for solving problem quickly" can be an indicator of user satisfaction.

The framework consists of three key modules: \textbf{Strategy Planner}, \textbf{Feature Retriever}, and \textbf{Score Analyzer} (Figure \ref{fig:framework}). The Strategy Planner generates interpretable natural language strategies for classifying user satisfaction by leveraging effective and ineffective strategies stored in memory. The Feature Retriever quantifies the relevance between user utterances and generated strategies, converting the LLM knowledge into measurable features for analysis. The Score Analyzer evaluates the effectiveness of strategies and categorizes them according to their contributions to USE improvement. Through iterative refinement of this process, PRAISE aims to identify optimal strategies to maximize overall performance.

PRAISE achieves state-of-the-art performance on USE benchmark datasets like MWOZ, SGD, and ReDial, while providing interpretable results. Beyond its high performance and interpretability, PRAISE brings additional advantages, including efficient inference and scalable deployment. The key contributions of our paper are:

\begin{itemize}
\item We propose PRAISE, a novel LLM-powered framework that automates the generation and validation of strategies specifically designed for User Satisfaction Estimation.
\item We present an approach that offers utterance-level interpretability, providing crucial insights for improving dialogue systems.
\item We transform LLM knowledge into measurable features, enabling efficient and scalable inference without direct LLM usage.
\item We demonstrate the effectiveness of PRAISE through extensive experiments on benchmark datasets, showing its robustness.
\end{itemize}

\section{Problem statement}
Consider a dataset $D$ of conversations, where each conversation $C_{i}$ is represented as a sequence of $T$ utterance pairs between the user and the assistant, denoted as $C_{i} = [(U_{i,1}, A_{i,1}),..., (U_{i,T}, A_{i,T})]$. Here, $U_{i,n}$ represents the n-th user utterance in the i-th conversation, and $A_{i,n}$ represents the corresponding assistant response. Each user utterance $U_{i, n}$ is associated with a satisfaction label $y_{i, n}$, which can take one of three values: SAT, DSAT, or NEU. The user satisfaction estimation problem can be formulated as learning a satisfaction estimator $\mathcal{E}$ that maps the current user utterance $U_{i,n}$ and the preceding conversation context to the corresponding satisfaction label $y_{i,n}$.

\vspace{-1.5em}
\begin{flalign*}
&\textbf{Satisfaction estimator (} \mathcal{E} \textbf{):}& \\ 
&\{(U_{i,1}, A_{i,1}),...,(U_{i,n-1}, A_{i,n-1}), U_{i,n}\} \rightarrow y_{i,n}&
\end{flalign*}

\section{Method}

\begin{figure*}[!htbp]
    \vspace{-10pt}
    \centering
    \includegraphics[width=1\linewidth]{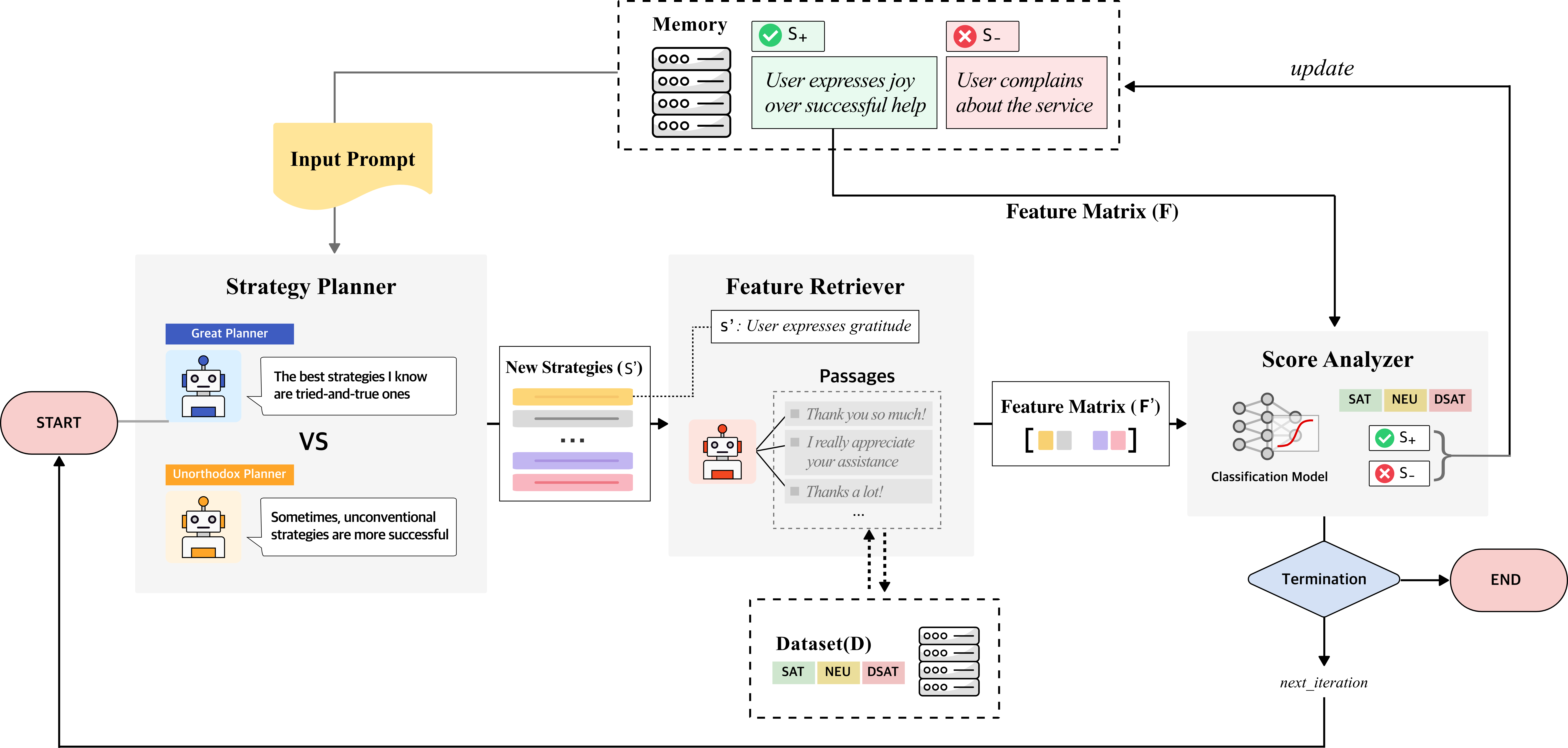}
    \caption{The overall framework of PRAISE.}
    \label{fig:framework}
\end{figure*}

\subsection{PRAISE: Plan and Retrieval Alignment for Interpretable Satisfaction Estimation}

The PRAISE framework is designed to address the challenges of interpretability and scalability in USE models. PRAISE consists of three key modules: \textit{Strategy Planner}, \textit{Feature Retriever}, and \textit{Score Analyzer}. The Strategy Planner uses LLMs to formulate hypothetical strategies for classifying user satisfaction. These strategies are interpretable natural language criteria that indicate user satisfaction levels in dialogues. The Feature Retriever generates passages based on these strategies and compares them with user utterances to extract quantified relevance features between the utterances and strategies. The Score Analyzer uses these features to compute a user satisfaction score, which is then used to classify user satisfaction levels. As a result, strategies that enhance classification performance are incorporated into effective strategies ($S_{\scriptscriptstyle+}$), while others go into ineffective strategies ($S_{\scriptscriptstyle-}$).

\subsubsection{Strategy Planner}

The Strategy Planner generates strategies for classifying user satisfaction using LLMs. It requires a problem-defining prompt and two types of strategies from previous USE evaluations: effective strategies ($S_{\scriptscriptstyle+}$) and ineffective strategies ($S_{\scriptscriptstyle-}$). These $S{\scriptscriptstyle+}$ and $S{\scriptscriptstyle-}$ guide the generation of new strategies to optimize overall performance, enhancing reasoning ability and avoiding redundancy. The planner then produces $n_{s}$ strategies that define the scenarios for SAT, DSAT, and NEU. Initially, 3 to 5 human-defined strategies are provided as $S_{\scriptscriptstyle+}$. Through iterations, $S_{\scriptscriptstyle+}$ and $S_{\scriptscriptstyle-}$ are updated based on improvements in user satisfaction classification.

We employ two distinct planner types: the \textbf{Great Planner} and the \textbf{Unorthodox Planner}. The great planner, operating at a lower temperature, generates strategies directly relevant to user satisfaction analysis. However, the great planner tends to generate consistent strategies when there are only minor changes in $S_{\scriptscriptstyle+}$ and $S_{\scriptscriptstyle-}$. To address this, we introduce the unorthodox planner which not only operates at a higher temperature but also receives specific prompts encouraging the generation of unconventional yet plausible strategies. The framework selects between these planners based on an exploration ratio ($\epsilon$), choosing the unorthodox planner with this probability. If the validation score does not improve, the exploration ratio doubles to expand the search space. When improvement occurs, the ratio resets to its initial value $\epsilon$.

\subsubsection{Feature Retriever}
\label{retriever}

The feature retriever quantifies the relationship between user utterances and strategies previously generated by the planner, aiming to identify the user utterances that best align with each strategy. To achieve this, it calculates the similarity between each strategy and individual user utterances. For effective similarity calculation, we adopt the retrieval method proposed by \citet{hyde}. The entire retrieval process consists of two stages: passage generation and feature retrieval.

\paragraph{Passage Generation}
In the first stage, the feature retriever generates $k$ hypothetical user passage examples $p_{s^{\prime}} = \{p_{s^{\prime},1}, p_{s^{\prime},2},..., p_{s^{\prime},k}\}$ that correspond to each strategy $s^{\prime}$ produced by the planner. As illustrated in the feature retriever example in Figure \ref{fig:framework}, for a strategy like "User expresses gratitude", the passage generation step could produce examples such as "Thank you so much!", or "I really appreciate your help". This process creates plausible user passages for each strategy $s^{\prime}$. Subsequently, these passages are transformed into embeddings $E_{p_{s^{\prime}}} \in \mathbb{R}^{k \times d}$ by an embedding model, where $d$ represents the embedding dimension.
The generation of multiple passages for each strategy ensures capturing a diverse range of potential user expressions, enabling more accurate similarity calculations between strategies and actual user utterances.

\paragraph{Feature Retrieval}
To retrieve relevant features for each strategy, we compute the matrix product between the generated passage embeddings $E_{p_{s^{\prime}}}$ and the embeddings of actual user utterances from the dataset, denoted as $E_u \in \mathbb{R}^{m \times d}$, where $m$ is the number of utterances in the dataset. This step is crucial as it allows us to measure the similarity between our hypothetical strategy-based passages and the real user responses.
The matrix product operation generates a relevance matrix $\mathcal{R} \in \mathbb{R}^{k \times m}$ for all combinations of generated passages and actual utterances. Each element in this matrix represents the similarity score between a generated passage and a real utterance. We then sum these relevance scores across all passages to obtain a single feature score for each strategy $s^{\prime}$:
$$f^{\prime}_{s^{\prime}} = \sum_{i=1}^k \mathcal{R}_i$$

The term $f^{\prime}_{s^{\prime}} \in \mathbb{R}^{m}$ represents the overall relevance score between a generated strategy ($s^{\prime}$) and each utterance. We then stack these scores for all newly generated strategies to form a feature matrix $F^{\prime} \in \mathbb{R}^{m \times n_{s}}$, where $n_{s}$ is the number of newly generated strategies. By quantifying strategies as similarity scores for each utterance, we create structured data that conventional machine learning models can effectively process and analyze.

\subsubsection{Score Analyzer}
\label{analyzer}

The score analyzer evaluates the effectiveness of generated strategies and trains the model to classify user satisfaction level. This module refines the strategy set and improves the overall performance of the user satisfaction.

We compute a baseline score ($\text{score}_0$) using the feature matrix $F$ from the previous best strategies ($S{\scriptscriptstyle +}$). We employ logistic regression as the final classification model ($\mathcal{M}$) and evaluate its performance on the validation set. We add new features ($F^{\prime}$) to the existing ones ($F$) column by column and evaluate using $\mathcal{M}$ to compare the resulting score with $\text{score}_0$. Strategies that make the score better go into $S^{\prime}{\scriptscriptstyle +}$, others into $S^{\prime}{\scriptscriptstyle -}$. We then update our overall strategy sets $S_{\scriptscriptstyle +}$ and $S_{\scriptscriptstyle -}$. 

However, continuously increasing the number of strategies can lead to longer prompts and reduced interpretability.  To solve this issue, we implement top-k strategy selection, which identifies and uses only the most useful strategies as effective ($S_{\scriptscriptstyle +}$) to optimize the combination. In the logistic regression model, the absolute values of the coefficients for all labels are summed, and the top-k features with the largest values are selected. When using other models, model-specific importance measures \cite{rf} or model-agnostic importance calculation \cite{shap, lime} can be employed. The pseudo-code of the score analyzer is shown in \textbf{algorithm \ref{alg:feature_addition}}.

\begin{algorithm}
\caption{\small{Selective Feature Addition Based on Score Improvement with Top-k Selection}}
\label{alg:feature_addition}
\scriptsize
\begin{algorithmic}
\setlength{\baselineskip}{0.85\baselineskip}
\State \( S^{\prime}_{\scriptscriptstyle +}, S^{\prime}_{\scriptscriptstyle -} \gets \emptyset \) \Comment{Initialize strategies set}
\State $ length \gets \Call{ColumnCount}{ F^\prime }$ \Comment{Get columns in $F^\prime$}
\State $ \text{score}_0 \gets \Call{Evaluate}{ \mathcal{M}, F}$ \Comment{Evaluate original features}
\For{\( j = 0 \) to \( length \)}
    \State \( \text{score} \gets \Call{Evaluate}{ \mathcal{M}, F \oplus F^\prime[:, j]} \)
    \State \( \text{improvement} \gets \text{score} - \text{score}_0 \)
    \If{\( \text{improvement} > 0 \)}
        \State \( S^{\prime}_{\scriptscriptstyle +} \gets S^{\prime}_{\scriptscriptstyle +} \cup \{S^{\prime}[j]\} \)
    \Else
        \State \( S^{\prime}_{\scriptscriptstyle -} \gets S^{\prime}_{\scriptscriptstyle -} \cup \{S^{\prime}[j]\} \)
    \EndIf
\EndFor
\State \( \text{coef} \gets \Call{GetCoefficients}{\mathcal{M}} \) \Comment{Get coefficients for all classes}
\State Let $L$ be the number of classes in the classification problem
\State \( \text{importance}_i = \sum_{l=1}^L |\text{coef}_{i,l}|, \quad \forall i \in \{1,\ldots,|S^{\prime}_{\scriptscriptstyle +}|\} \)
\State Sort \( S^{\prime}_{\scriptscriptstyle +} \) in descending order based on \( \text{importance} \)
\State \( S^{\prime}_{\scriptscriptstyle +} \gets \text{first } k \text{ elements of sorted } S^{\prime}_{\scriptscriptstyle +} \)
\State \textbf{return} \( S^{\prime}_{\scriptscriptstyle +}, S^{\prime}_{\scriptscriptstyle -} \)
\end{algorithmic}
\end{algorithm}

\subsection{Inference}

During the inference phase, PRAISE employs its trained model and refined strategies to classify user satisfaction for individual utterances in new dialogues. The feature retriever operates similarly to the training phase but uses only the final set of effective strategies ($S_{\scriptscriptstyle+}$). Specifically, the feature vector $F$ is calculated using the passage embeddings $E_p$ from the training phase and the new utterance embeddings $E_u^{test}$. This approach of reusing the passages has two key advantages: it ensures consistency with the optimal results and reduces computational costs by eliminating the need for additional passage generation. This feature vector is then passed to the score analyzer, where the logistic regression model predicts the user satisfaction level for the utterance. This approach enables PRAISE to evaluate user satisfaction efficiently without LLM inference, ensuring scalability through simple models. Additionally, the feature vector $F$ and $S_{\scriptscriptstyle+}$ provide utterance-level interpretability.

\section{Experiments}
\subsection{Experimental setup}
\label{experimental_setup}

\renewcommand{\arraystretch}{1.5}
\begin{table}[!htbp]
\centering
\resizebox{\columnwidth}{!}{%
\footnotesize
\setlength{\tabcolsep}{4pt}
\begin{tabular}{lccc}
\thickhline
 \textbf{Dataset} & \textbf{MWOZ} & \textbf{SGD} & \textbf{ReDial} \\
\hline
\# Conversations & 1,000 & 1,000 & 1,000 \\
\# Utterances & 21,706 & 24,148 & 16,616 \\
\# Labeled utterances & 8,439 & 9,316 & 7,304 \\\hline
\makecell[l]{Label distribution \\ (SAT / NEU / DSAT, \%)} & 40.4 / 32.2 / 27.4 & 47.6 / 30.2 / 22.2 & 49.5 / 26.9 / 23.6 \\
\thickhline
\end{tabular}
}
\caption{\footnotesize Statistics of datasets.}
\label{data_stat}
\end{table}

\renewcommand{\arraystretch}{1.2}
\setlength\dashlinedash{0.5pt}
\setlength\dashlinegap{1.5pt}
\setlength{\tabcolsep}{11pt}
\begin{table*}[t]
\centering
\resizebox{\textwidth}{!}{%
\begin{tabular}{cccccccccccccc}
\thickhline
\multirow{2}{*}{\textbf{Model}} & \multicolumn{4}{c}{\textbf{MWOZ}} & \multicolumn{4}{c}{\textbf{SGD}} & \multicolumn{4}{c}{\textbf{ReDial}} \\ \cline{2-13}
 & \textbf{Acc} & \textbf{Prec} & \textbf{Rec} & \textbf{F1} & \textbf{Acc} & \textbf{Prec} & \textbf{Rec} & \textbf{F1} & \textbf{Acc} & \textbf{Prec} & \textbf{Rec} & \textbf{F1} \\ \hline
HiGRU \cite{higru} & 44.6 & 43.7 & 44.3 & 43.7 & 50.0 & 47.3 & 48.4 & 47.5 & 46.1 & 44.4 & 44.0 & 43.5 \\ \hdashline
BERT \cite{bert} & 46.1 & 45.5 & 47.4 & 45.9 & 56.2 & 55.0 & 53.7 & 53.7 & 53.6 & 50.5 & 51.3 & 50.0 \\ \hdashline
USDA \cite{usda} & 49.9 & 49.2 & 49.0 & 48.9 & 61.4 & 60.1 & 55.7 & 57.0 & 57.3 & 54.3 & 52.9 & 53.4 \\ \hdashline
ASAP \cite{asap}& 56.6 & 55.1 & 54.9 & 54.9 & 64.4 & \underline{62.7} & 62.5 & \underline{62.5} & 62.9 & 60.2 & 60.4 & 60.0 \\ \hline
GPT-3.5-turbo & 36.3 & \underline{57.3} & 43.9 & 30.9 & 51.6 & 40.9 & 43.3 & 37.1 & 55.1 & 47.9 & 44.3 & 39.7 \\ \hdashline
(+ 3-shots) & 42.2 & 42.2 & 43.0 & 41.6 & 50.3 & 48.4 & 48.0 & 47.7 & 54.4 & 53.8 & 47.1 & 46.5 \\ \hdashline
GPT-4 (+ 3-shots) & 45.8 & 46.0 & 45.4 & 43.8 & 52.8 & 48.7 & 43.3 & 42.7 & 53.5 & 51.0 & 45.4 & 45.1 \\ \hdashline
SPUR \cite{spur}& 49.8 & \textbf{67.1} & 40.6 & 36.4 & 48.4 & 40.8 & 36.2 & 32.1 & 56.1 & 56.3 & 47.3 & 42.3 \\ \hline
\textbf{PRAISE (30)} & \textbf{60.3} & 55.8 & \textbf{59.9} & \underline{56.7} & \underline{64.5} & 61.6 & \underline{63.0} & 62.2 & \textbf{66.0} & \underline{64.4} & \underline{63.6} & \underline{63.8} \\ \hdashline
\textbf{PRAISE (50)} & \textbf{60.3} & 56.6 & \underline{59.8} & \textbf{57.3} & \textbf{65.6} & \textbf{63.0} & \textbf{63.6} & \textbf{63.2} & \textbf{66.0} & \textbf{64.5} & \textbf{63.7} & \textbf{63.9} \\ \thickhline
\end{tabular}%
}
\caption{Performance of models on MWOZ, SGD, and ReDial datasets. Numbers in parentheses after PRAISE indicate the maximum number of features used. \textbf{Bold} is the best performance, while \underline{underlined} is the second-best.}
\label{result}
\end{table*}

\paragraph{Datasets and metrics}
We use three task-oriented dialogue datasets \cite{li2016user, wu2019global}: ReDial \cite{redial_data} for movie recommendations, and SGD \cite{sgd_data} and MWOZ \cite{mwoz_data} for general scenarios such as bookings and information requests. 
User satisfaction is annotated as `SAT' when the system effectively resolves user queries and achieves their goals, `DSAT' when it fails to meet user needs or provides irrelevant responses, and `NEU' when it partially fulfills the request \cite{uss_dataset}.
We have converted the original five-level ratings (1-5) to three satisfaction classes: DSAT (average rating < 3), NEU (average rating = 3), and SAT (average rating > 3), which is more practical for real-world applications and aligns with previous studies \cite{usda, asap}. The dataset is split into train, validation, and test sets with an 8:1:1 ratio, excluding dialogues fewer than two turns. Table \ref{data_stat} presents the dataset statistics. We use the same evaluation metrics as in previous works \cite{usda, asap}, including Accuracy (Acc), macro-averaged Precision (P), Recall (R), and F1-score (F1).

\paragraph{Baselines}
We compare our proposed method with the following baseline models:
\begin{enumerate}[label={-}, leftmargin=*, itemsep=0pt, topsep=0pt, parsep=2pt, partopsep=0pt]
\item \textbf{HiGRU} \cite{higru} utilizes two Bidirectional GRUs \cite{gru} structures to capture user utterances and context information.
\item \textbf{BERT} \cite{bert} estimates user satisfaction by taking the concatenation of previous dialogue context and the user's last utterance, separated by a [SEP] token, as input.
\item \textbf{USDA} \cite{usda} pre-trains on dialogue patterns leading to satisfaction using pseudo-labels, then employs an Attentive GRU model to estimate user satisfaction.
\item \textbf{ASAP} \cite{asap} combines BERT with a Hawkes process \cite{hawkes1, hawkes2} to better capture the temporal dynamics of the conversation.
\item \textbf{GPT family} includes GPT-3.5\footnote{https://platform.openai.com/docs/models\#gpt-3-5-turbo} and GPT-4 \cite{gpt4} models, which evaluate satisfaction based on instructional prompts in various settings such as zero-shot and few-shot learning.
\item \textbf{SPUR} \cite{spur} uses iterative prompting with LLMs to generate rubrics from conversations, which are then used to evaluate user satisfaction.
\end{enumerate}

The baseline SPUR model only predicts SAT/DSAT, excluding NEU. Therefore, we define a range for the satisfaction score ($-k\sim k$) and predict the user satisfaction as NEU if it falls within that range. We set the $k$-value that maximizes the F1-score for each validation set.

\paragraph{Implementation details}
PRAISE implements a strategy planner using GPT-4 (gpt-4-1106-preview) for generating strategies and a feature retriever employing GPT-3.5-turbo (gpt-3.5-turbo-0125) for passage generation. Both modules handle five strategies ($n_s$) and passages ($k$) respectively. The exploration ratio ($\epsilon$) is set to 0.1. For text embedding, we use OpenAI's text-embedding-3-large\footnote{https://platform.openai.com/docs/guides/embeddings} model with 1024 dimensions. The training process involves 50 iterations using macro-F1, with early stopping if the validation score fails to improve for five consecutive iterations. For the logistic regression component, we employed l2 penalty with C=100. To ensure convergence, max\_iter was set to 500 for MWOZ and 700 for ReDial and SGD. All main experiments were conducted using NVIDIA H100 (80GB), and for inference speed experiments in Section \ref{inference_time}, we additionally used GTX 1080Ti (11GB).

\begin{figure*}[t]
\begin{adjustbox}{width=1.0\textwidth,center}
\begin{minipage}[t]{0.6\textwidth}
    \vspace{0pt} 
    \centering
    \includegraphics[width=\textwidth]{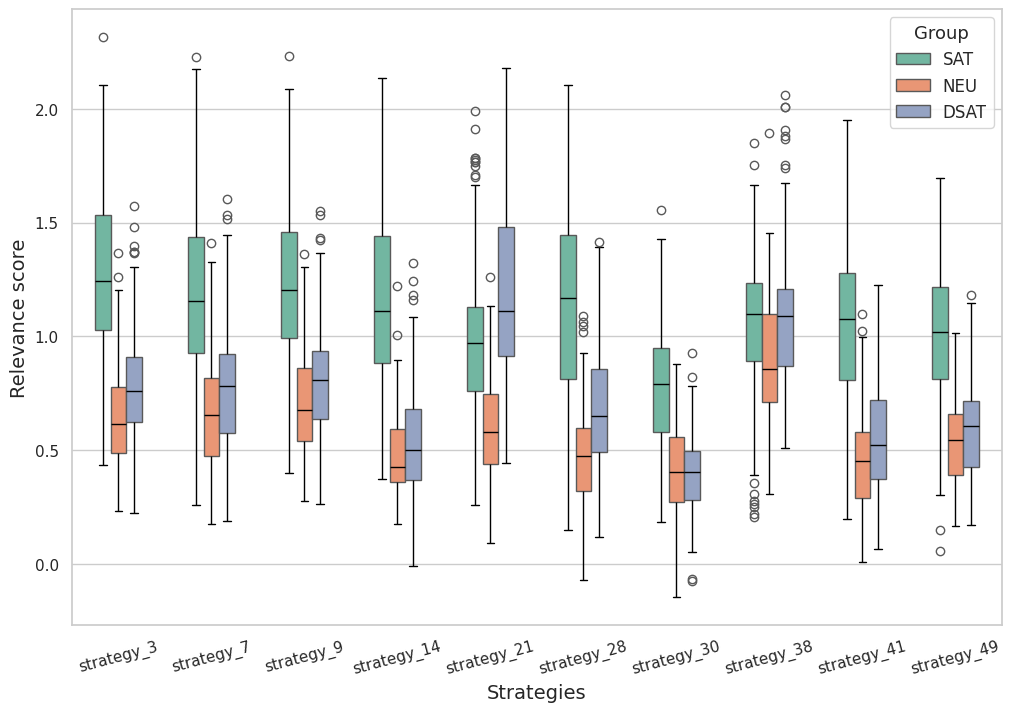}
\end{minipage}%
\hfill
\begin{minipage}[t]{0.4\textwidth}
    \vspace{14pt} 
    \tiny 
    \begin{itemize}[leftmargin=*]
        \item \textbf{strategy\_3 : User is thankful for the alternative suggestion.}
        \item strategy\_7 : User expresses appreciation for assistance.
        \item strategy\_9 : User thanks for quick response.
        \item strategy\_14 : User acknowledges helpful recommendation.
        \item \textbf{strategy\_21 : User declines with "No".}
        \item strategy\_28 : User indicates problem was resolved.
        \item strategy\_30 : User praises the assistant's efficiency.
        \item strategy\_38 : User requests additional information.
        \item strategy\_41 : User expresses relief after receiving information.
        \item strategy\_49 : User highlights ease of process.
    \end{itemize}
\end{minipage}
\end{adjustbox}
\caption{Box-plot of relevance scores for strategies in the SGD dataset.}
\label{fig:user_feedback_analysis}
\end{figure*}

\subsection{Main Results}
\begin{figure*}[!htbp]
\centering
\begin{minipage}[t]{\textwidth}
    \includegraphics[width=1\linewidth]{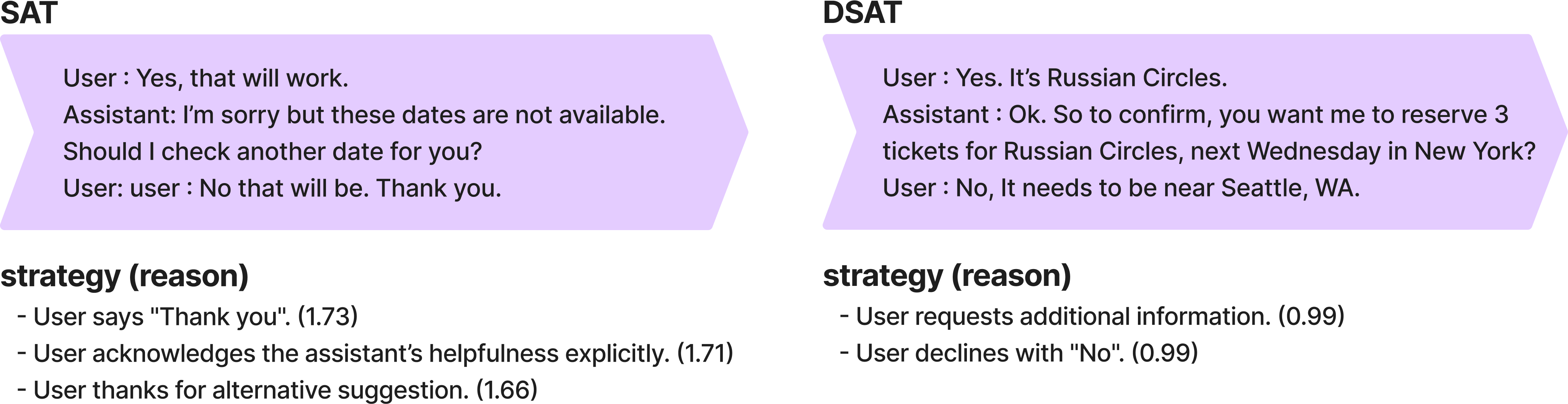}
    \caption{Strategies as interpretable reasons for predicting satisfaction.}
    \label{fig:some_sample}
\end{minipage}
\end{figure*}

Table \ref{result} demonstrates that PRAISE achieves state-of-the-art performance across most metrics and datasets in user satisfaction estimation. PRAISE obtains the highest F1-scores of 57.3\%, 63.2\%, and 63.9\% on the MWOZ, SGD, and ReDial datasets, respectively. Additionally, it achieves the best accuracy scores of 60.3\%, 65.6\%, and 66.0\% on MWOZ, SGD, and ReDial, respectively. Among the baselines utilizing Large Language Models (LLMs), including GPT-3.5-turbo, GPT-4, and SPUR, performance is relatively poor even with 3-shot learning. This suggests that direct application of LLMs, without task-specific fine-tuning or adaptation, may not be sufficient for accurate user satisfaction estimation. Notably, SPUR shows suboptimal performance despite being the most recent approach. This indicates that evaluating user satisfaction across entire conversation sessions may not be well-suited for utterance-level satisfaction assessment. Among the baseline models, ASAP demonstrates the best performance, though still not surpassing PRAISE.


\subsection{Interpretability}
The feature matrix $F$ contains relevance scores that show how each utterance relates to the generated strategies. These scores provide interpretability by indicating which strategy is most relevant to a given utterance.

Figure \ref{fig:user_feedback_analysis} shows the relevance score distribution across satisfaction levels, illustrating that certain strategies effectively distinguish between SAT and DSAT. For example, "User is thankful for the alternative suggestion." (strategy\_3) and "User declines with `No'." (strategy\_21) exhibit higher values for SAT and DSAT, respectively. These strategies provide clear explanations on distinguishing satisfaction levels in the current dataset.

The relevance scores for individual utterances provide insights into the reasons behind the predicted satisfaction level. Figure \ref{fig:some_sample} demonstrates how these scores of the last utterance in a conversation can explain why it was classified as SAT, NEU, or DSAT. Additional examples are provided in Appendix \ref{apdx:interpretability_b}.

\subsection{Scalability}
\label{inference_time}

\begin{figure}[h]
    \centering
    \resizebox{\columnwidth}{!}{\includegraphics{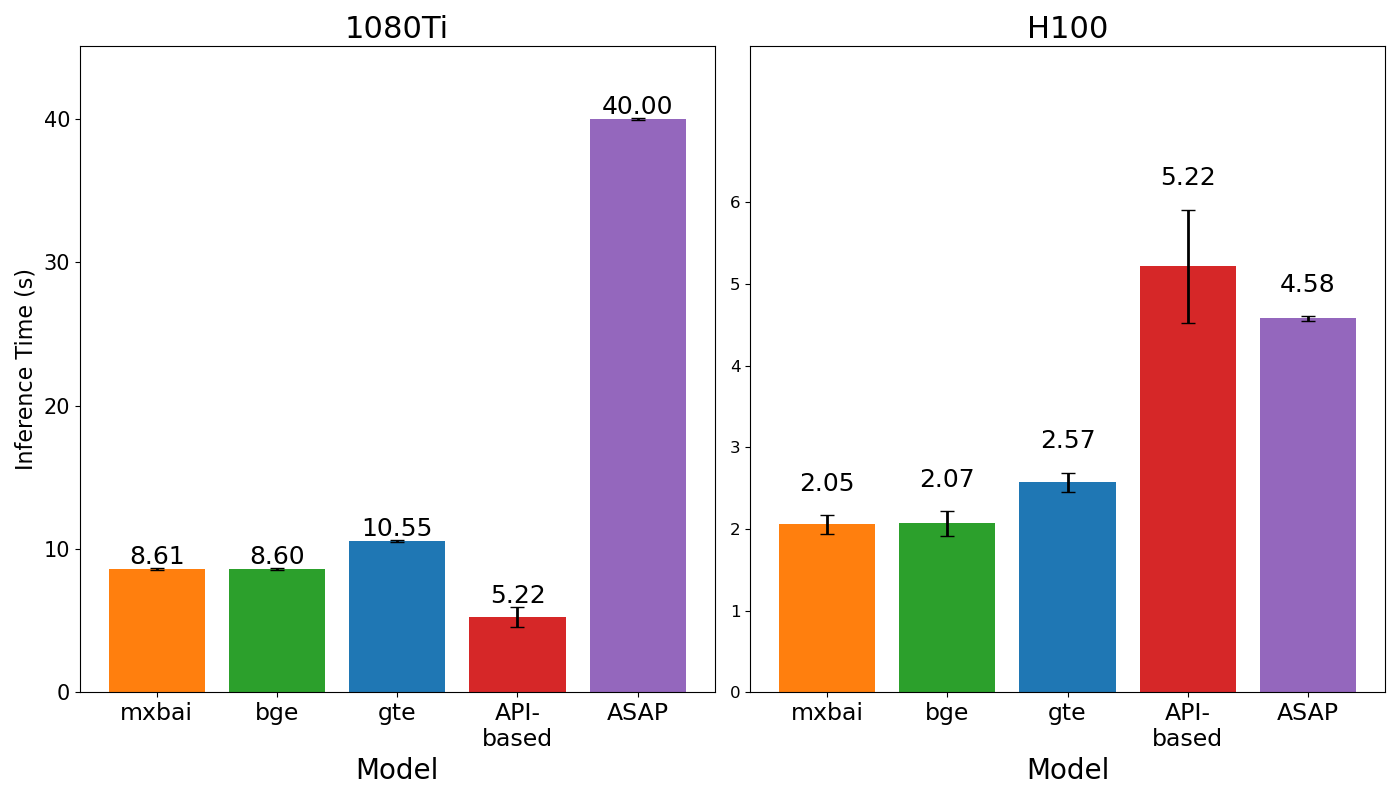}}
    \caption{Comparison of inference time between PRAISE with various embedding models and ASAP}
    \label{fig:inference_time}
\end{figure}

In this study, we evaluate the scalability of the PRAISE model by comparing its inference time with the current state-of-the-art model, ASAP. A shorter inference time indicates better scalability, as it implies efficient utilization of computational resources. To evaluate inference times across different GPU environments, we test these models on high-end (NVIDIA H100) and low-end (NVIDIA GTX 1080Ti) setups, repeating each experiment 30 times for reliability. GPT-based models are excluded due to their lower performance and significantly longer inference times. As shown in Figure \ref{fig:inference_time}, the ASAP model is slower than the others on the low-end GPU, while the API-based model (i.e., text-embedding-3-large) is the fastest. In the high-end GPU environment, ASAP runs faster than the API-based model but remains slower than the other models. This shows that PRAISE with API-based model are the most effective in lower-spec GPU environments. Even without API access, PRAISE with other models are more efficient than ASAP. Also, PRAISE demonstrates high scalability by eliminating the need for LLMs during inference, utilizing only a pre-trained logistic regression model, the passage embedding matrix $E_p$, and an embedding model.

\subsection{Ablation study} 

\setlength{\tabcolsep}{6pt} 
\renewcommand{\arraystretch}{1.1} 
\begin{table*}[!t]
\centering
\small
\resizebox{\textwidth}{!}{%
\begin{tabular}{@{}ccccccccccccc@{}}
\toprule
\raisebox{0.0em}{\multirow{2}{*}{\parbox[c]{2cm}{\centering \textbf{Embedding \\ Model}}}} & \multicolumn{4}{c}{\textbf{MWOZ}} & \multicolumn{4}{c}{\textbf{SGD}} & \multicolumn{4}{c}{\textbf{ReDial}} \\ \cmidrule(lr){2-5} \cmidrule(lr){6-9} \cmidrule(lr){10-13} 
 & Acc & Prec & Rec & F1 & Acc & Prec & Rec & F1 & Acc & Prec & Rec & F1 \\ \midrule
mxbai-large \cite{mxbai} &  56.6 & 56.9 & 52.4 & 53.2 & 61.0 & 59.2 & 60.9 & 59.6 & 64.3 & 62.1 & 63.0 & 62.4 \\
bge-large-en \cite{bge} & 57.2 & 56.7 & 52.7 & 53.3 & 61.7 & \underline{59.9} & \underline{61.6} & \underline{60.3} & \underline{64.7} & \textbf{62.5} & \textbf{63.5} & \textbf{62.8} \\
gte-large-en \cite{gte} & \underline{57.9} & \underline{57.7} & \underline{54.2} & \underline{54.9} & \underline{61.9} & 59.8 & 58.6 & 59.0 & 64.3 & 62.3 & \underline{63.2} & 62.4 \\
\textbf{text-embedding-3-large} & \textbf{60.3} & \textbf{56.6} & \textbf{59.8} & \textbf{57.3} & \textbf{65.6} & \textbf{63.0} & \textbf{63.6} & \textbf{63.2} & \textbf{66.0} & \underline{64.5} & 63.7 & \underline{63.9} \\
\bottomrule
\end{tabular}
}
\caption{Satisfaction classification performance with different embedding models}
\label{tab:performance_comparison}
\end{table*}

We conducted an ablation study on the techniques used in each module. We performed 20 repetitions of the experiment for each setting, starting from the initial strategies.

\begin{figure*}[h]
\begin{adjustbox}{width=0.99\textwidth, center}
\centering
\begin{subfigure}{0.33\textwidth}
\centering
\includegraphics[width=\linewidth]{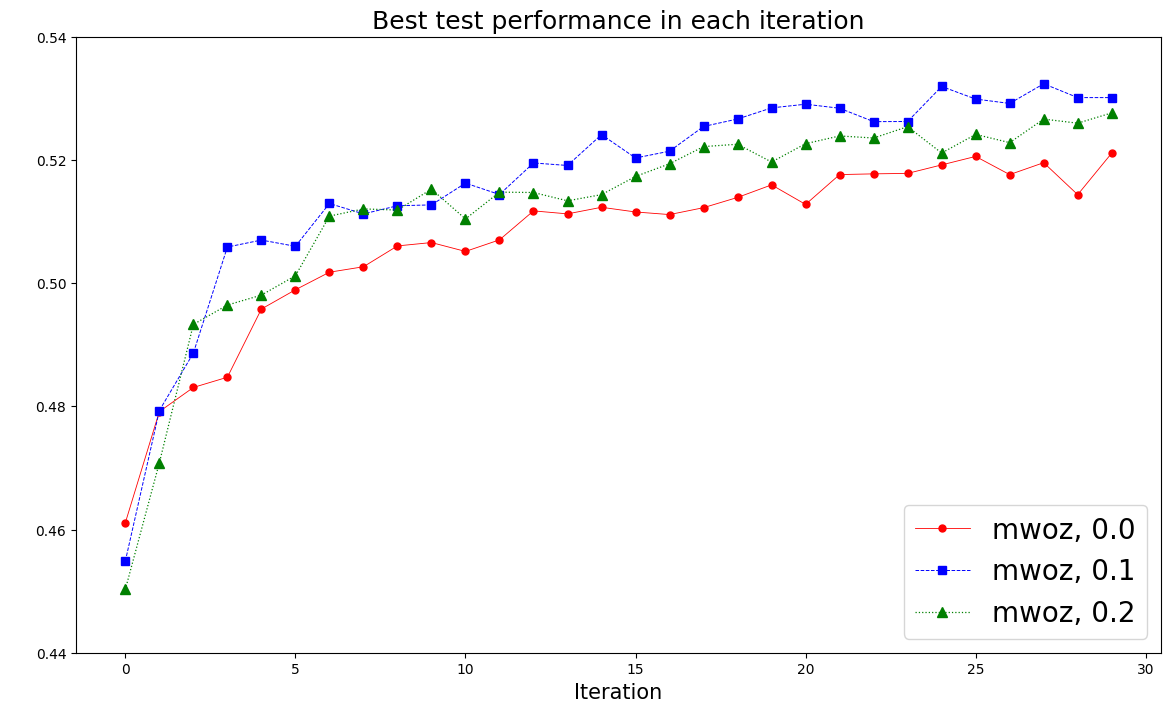}
\caption*{(a) MWOZ dataset}
\label{fig:sub1}
\end{subfigure}%
\begin{subfigure}{0.33\textwidth}
\centering
\includegraphics[width=\linewidth]{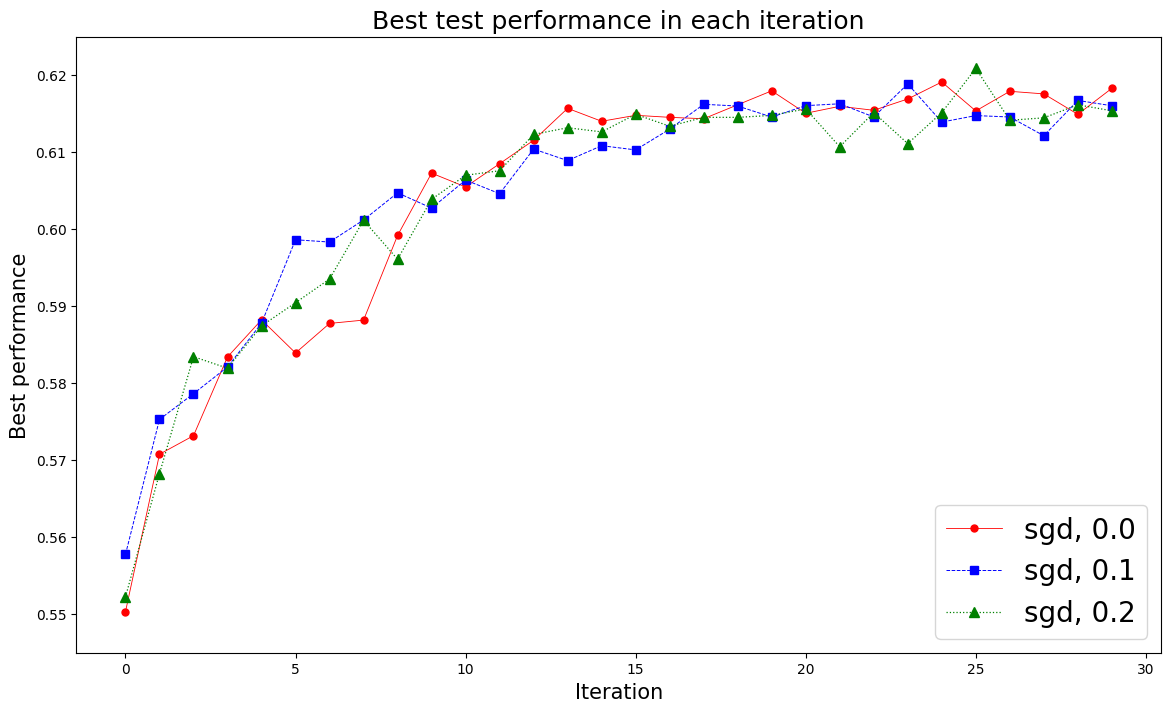}
\caption*{(b) SGD dataset}
\label{fig:sub2}
\end{subfigure}
\begin{subfigure}{0.33\textwidth}
\centering
\includegraphics[width=\linewidth]{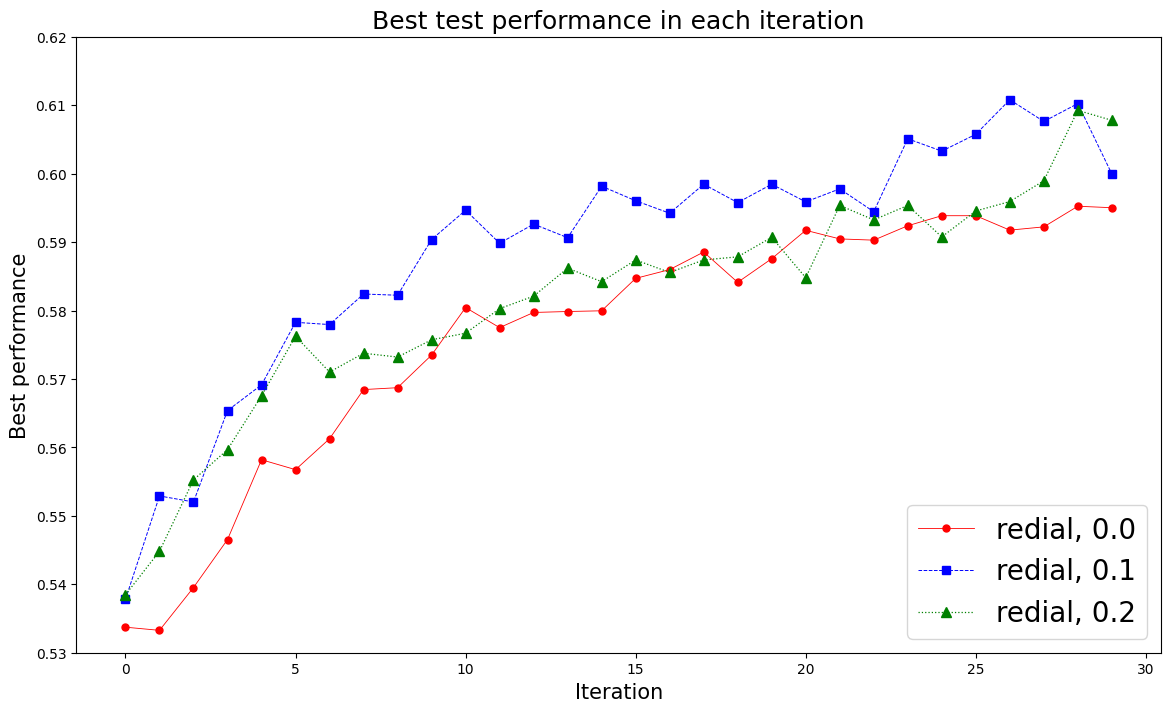}
\caption*{(c) ReDial dataset}
\label{fig:sub3}
\end{subfigure}
\end{adjustbox}
\caption{Effect of exploration ratio on maximum test score across iterations. We experimented with exploration ratios of 0.0, 0.1, and 0.2, where an exploration ratio of 0.0 means not using the unorthodox planner at all.}
\label{fig:max_plot}
\end{figure*}

\paragraph{Unorthodox planner} The probability of using the unorthodox planner is determined by the exploration ratio ($\epsilon$). This ratio improves the final results by generating strategies that the great planner cannot conceive. Figure \ref{fig:max_plot} shows that a 0.1 exploration ratio achieved higher maximum F1 scores in ReDial and MWOZ datasets. The SGD dataset did not show significant differences across various exploration ratios.

\paragraph{Embedding model} We tested PRAISE with various embedding models to verify its robustness and adaptability. Our results are consistent across different embeddings (Table \ref{tab:performance_comparison}). The text-embedding-3-large model we employ in PRAISE exhibited the best performance among the embeddings tested. Notably, other embedding models we tested also delivered competitive results compared to text-embedding-3-large. These findings suggest that the inference stage could potentially eliminate the reliance on API-based embedding models, leading to significantly more cost-effective and time-efficient model operation.

\paragraph{Passage generation} Our study utilizes a retrieval process based on passage generation inspired by \cite{hyde}. This approach performed significantly better than using embeddings alone (Table \ref{not_example}).

\renewcommand{\arraystretch}{1.1}
\begin{table}[!htbp]
\centering
\resizebox{\columnwidth}{!}{%
\scriptsize
\setlength{\tabcolsep}{4pt}
\begin{tabular}{@{}ccccccccc@{}}
\toprule
\multirow{2}{*}{\textbf{Model}} & \multicolumn{2}{c}{\textbf{MWOZ}} & & \multicolumn{2}{c}{\textbf{SGD}} & & \multicolumn{2}{c}{\textbf{ReDial}} \\
\cmidrule(lr){2-3} \cmidrule(lr){5-6} \cmidrule(lr){8-9}
& Acc & F1 & & Acc & F1 & & Acc & F1 \\
\midrule
Not Example & 58.3 & 55.6 & & 63.1 & 60.8 & & 63.2 & 61.2 \\
\textbf{PRAISE (50)} & \textbf{60.3} & \textbf{57.3} & & \textbf{65.6} & \textbf{63.2} & & \textbf{66.0} & \textbf{63.9} \\
\bottomrule
\end{tabular}
}
\caption{Satisfaction classification performance: with vs. without passage generation}
\label{not_example}
\end{table}

\paragraph{Impact of initial strategy selection on performance}

We investigated whether initial strategies influence subsequent performance. Although these strategies become less impactful throughout the training process, we considered it crucial to verify their potential long-term effects on the model's performance. We conducted an experiment using 5 sets of 5 randomly generated strategies for satisfaction classification, each trained 20 times over 50 steps. Our analysis, involving normality tests, homogeneity of variance tests, and ANOVA, revealed significant differences in average performance during the initial steps. However, these differences disappeared starting in the fourth iteration, with consistently high p-values in subsequent iterations. This finding suggests that while initial strategies influence performance in the early training stages, their impact diminishes as the PRAISE process continues. Further details on performance comparisons across different sets and steps can be found in Appendix \ref{multiple_run}.

\section{Related work}
\paragraph{User satisfaction estimation}

Prior works on user satisfaction estimation in dialogue systems have evolved from content-based methods, such as sentiment analysis \cite{song-etal-2019} and interaction quality assessment \cite{SCHMITT2015}, to leveraging pre-trained language models \cite{kachuee-etal-2021-self} and dialogue action tasks \cite{usda}, modeling satisfaction dynamics \cite{asap}, and using multi-task adversarial method \cite{song-23}. Recent efforts have explored using LLMs as simulators \cite{hu2023} and augmenting datasets with counterfactual dialogue samples \cite{abolghasemi2024cause}. Despite recent advancements, many of these methodologies still struggle with interpretability. To address these limitations, \citet{spur} proposed a framework using LLM, but it falls short in providing instance-level interpretability and incurs high costs when summarizing and classifying data using GPT-4. PRAISE achieves high efficiency by using LLM only to generate strategies during training, without requiring it for inference. Additionally, this approach provides interpretability for overall USE classification and utterance-level analysis.

\paragraph{Retrieval models}
Text embeddings that capture semantic similarity and context have been developed, ranging from early models \cite{liu2019roberta, mikolov2013efficient,pennington2014glove,reimers2019sentencebert} to larger models \cite{behnamghader2024llm2vec} trained on large-scale corpora. Diverse retrieval tasks utilize these embeddings, such as DPR \cite{karpukhin2020dense}, Contriever \cite{izacard2021unsupervised}, ANCE \cite{xiong2020approximate}, and RocketQA \cite{qu2020rocketqa}. However, applying these retrievers to domain-specific data is challenging due to the lack of relevance supervision data. To address this issue, studies have utilized LLMs to generate pseudo-labels, such as hypothetical documents \cite{gao2022precise} or generated relevant queries \cite{bonifacio2022inpars,boytsov2023inpars,dai2022promptagator,jeronymo2023inpars} for model training. To resolve the scarcity of relevance data, we employ LLMs to generate passages that align with the generated strategies. We then utilize the search results between these generated passages and actual user utterances as features in our framework.

\section{Conclusion}
\label{conclusion}

In this paper, we presented PRAISE, a novel framework for User Satisfaction Estimation (USE) in dialogue systems that leverages Large Language Models (LLMs). Our work addresses the crucial challenge of developing interpretable and scalable methods for assessing user satisfaction, a key factor in improving conversational AI. PRAISE consists of three main modules: the Strategy Planner, which generates natural language strategies for classifying user satisfaction; the Feature Retriever, which provides multi-level interpretability by aligning user utterances with strategies; and the Score Analyzer, which evaluates strategy effectiveness and enables efficient inference by transforming LLM knowledge into a structured representation. Our experimental results across three benchmark datasets demonstrate PRAISE's superior performance compared to existing USE methods. 

\section*{Limitations}
\label{limitation}

PRAISE has several limitations that require further exploration, which can be summarized into three main points:

\paragraph{Embedding models.}
In PRAISE, we use a basic embedding model without fine-tuning. Enhancing the embedding model to better understand complex dialogue contexts and user intent, specifically for the USE task, could improve performance.

\paragraph{LLM-Driven Strategies.}
As the effectiveness of PRAISE Strategies is heavily dependent on the internal knowledge of LLM, their performance might be significantly compromised in domains where the LLM has limited or insufficient information, potentially leading to suboptimal or unreliable outputs. Future work should integrate external knowledge or additional modules to support strategy generation process.

\paragraph{Open-Domain Dialogues Evaluation.}
PRAISE currently focuses on task-oriented dialogue datasets, as these are the only datasets with user satisfaction annotations. In the future, evaluating PRAISE on open-domain dialogues, which more closely resemble real-world conversations, will be essential for expanding its practical applications.

\section*{Acknowledgements}
This work was partly supported by the BK21 FOUR Program (Education and Research Center for Industrial Innovation Analytics) funded by the Ministry of Education, Korea

\bibliography{custom}

\appendix
\section{Prompt}

\subsection{Problem definition}
\label{pd}
Table \ref{tab:problem_definition} shows the common prompts and problem definition of each dataset.

\captionsetup[table]{skip=7pt}
\begin{table*}[t]
\begin{tabularx}{\textwidth}{X}
\toprule
\textbf{Common prefix prompt} \\
\midrule
\fontsize{8.5pt}{9pt}\selectfont
You are a competent bot that generates strategies to classify conversations in which the user 
expresses satisfaction. \\
\midrule
\textbf{MWOZ} \\
\midrule
\fontsize{8.5pt}{9pt}\selectfont
The User and Assistant are having a conversation about making a reservation for a specific service, or looking up information such as an address or phone number.

The types of services include taxis, restaurants, buses, hotels, attractions, and trains.

The user asks a number of questions about the service, and their satisfaction depends on the assistant's answers.

Users are satisfied if the assistant answers their questions appropriately, but they are also dissatisfied if the service provider does not provide the information they asked for, regardless of the assistant's answer. \\
\midrule
\textbf{SGD} \\
\midrule
\fontsize{8.5pt}{9pt}\selectfont
Assistant is a virtual assistant that provides information about Alarm, Bank, 
transportation(bus, flight, etc.), reservation(rental car, restaurant etc.), Calendar, Event, Home, Hotel, Media, Movie, Music, Service, Travel, Weather and many other things people might want to know.

A typical satisfaction for a user is when they successfully make a reservation or find the assistant's suggestions helpful, and sometimes they are dissatisfied with the assistant's answer and ask for another alternative or decline.

Include specific context in your strategy for the information the assistant provides. (e.g. user requests a bus at a different time.) \\
\midrule
\textbf{ReDial} \\
\midrule
\fontsize{8.5pt}{9pt}\selectfont
The user and the assistant have a conversation about movies, talking about the movies they've seen or recommending movies to each other.

The Assistant's suggestions, questions, and reactions have a significant impact on the user's satisfaction, which can be inferred from the user's conversations.

The main topics of conversation are the title, actors, and genre of the movie, but they also include casual conversation. \\
\bottomrule
\end{tabularx}
\caption{Common prompts and problem definition of each dataset}
\label{tab:problem_definition}
\end{table*}

\subsection{Initial strategies}
\label{init_strategies}

Table \ref{tab:initial_strategies} shows the initial strategies for each dataset that we used in our experiment.

\begin{table}[h]
\small
\begin{tabular}{p{0.95\columnwidth}}
\hline
\textbf{MWOZ} \\
\hline
\begin{itemize}[leftmargin=*, nosep]
  \item The user thanks the assistant.
  \item The user repeats the same question.
  \item The user asks about other services.
\end{itemize} \\
\hline
\textbf{ReDial} \\
\hline
\begin{itemize}[leftmargin=*, nosep]
  \item User asks for more movie recommendations.
  \item User expresses interest in a movie's director.
  \item User compliments assistant's choice.
  \item User requests further details on movie.
  \item User expresses interest in a specific genre.
\end{itemize} \\
\hline
\textbf{SGD} \\
\hline
\begin{itemize}[leftmargin=*, nosep]
  \item User expresses satisfaction with the service quality.
  \item User acknowledges assistant's quick thinking.
  \item User shows appreciation for assistance.
  \item User empathizes with the assistant.
  \item User appreciates the detailed explanation.
\end{itemize} \\
\hline
\end{tabular}
\caption{Initial strategies of each dataset}
\label{tab:initial_strategies}
\end{table}

\subsection{Strategy planner prompts}
\label{planner_prompts}

In all prompts, \{\{\$variable\}\} acts as a placeholder to accept external variables.

For example, \{\{\$problem\_definition\}\} takes as input the problem definition for each dataset in Appendix \ref{pd}.

\subsubsection{Great planner}

\begin{Verbatim}[commandchars=\\\{\}, fontsize=\fontsize{8pt}{9pt}\selectfont]
\small{hyperparameter} : \{
  "model": "gpt-4-1106-preview",
  "temperature": 0.1,
  "max_tokens": 512
\}
\end{Verbatim}

\textbf{system prompt}
\begin{Verbatim}[commandchars=\\\{\}, fontsize=\fontsize{6pt}{9pt}\selectfont]
    \{\{$problem_definition\}\}

    [ouput format]
    Your answer should be in the following json format
    \{
      "strategies": [
        "User [common verb] [appropriate object less than 5 words].",
        "User [common verb] [appropriate object less than 5 words].",
        ...
      ]
    \}

    Below are the strategies created so far
    
    [Effective strategies]
    \{\{$effective_strategies\}\}
    
    [Ineffective strategies]
    \{\{$ineffective_strategies\}\}
    
    Generate \{\{$strategy_num\}\} additional effective strategies that
    you think would help your analysis.
    answer:
\end{Verbatim}

\subsubsection{Unorthodox planner}

\begin{Verbatim}[commandchars=\\\{\}, fontsize=\fontsize{8pt}{9pt}\selectfont]
\small{hyperparameter} : \{
  "model": "gpt-4-1106-preview",
  "temperature": 0.7,
  "max_tokens": 512
\}
\end{Verbatim}

\textbf{system prompt}
\begin{Verbatim}[commandchars=\\\{\}, fontsize=\fontsize{6pt}{9pt}\selectfont]
    [problem definition]
    \{\{$problem_definition\}\}

    [ouput format]
    Your answer should be in the following json format
    \{
      "strategies": [
        "User [common verb] [appropriate object that fits the strategy].",
        "User [common verb] [appropriate object that fits the strategy].",
        ...
      ]
    \}

    Below are the strategies created so far
    
    [Effective strategies]
    \{\{$effective_strategies\}\}
    
    [Ineffective strategies]
    \{\{$ineffective_strategies\}\}
    
    In our opinion, the above strategies are too formulaic,
    and sometimes crazy strategies that are completely weird
    or nonsensical are more successful.
        
    Generate \{\{$strategy_num\}\} strategies that sound like conversations
    you'd have in a problem definition situation, but don't seem to have
    anything to do with user satisfaction.
    
    answer:
\end{Verbatim}

\subsection{Feature retriever prompts}
\label{retriever_prompts}
\subsubsection{Passage generator}

\begin{Verbatim}[commandchars=\\\{\}, fontsize=\fontsize{8pt}{9pt}\selectfont]
\small{hyperparameter} : \{
  "model": "gpt-3.5-turbo-0125",
  "temperature": 0.0,
  "max_tokens": 1024
\}
\end{Verbatim}

\textbf{system prompt}
\begin{Verbatim}[commandchars=\\\{\}, fontsize=\fontsize{6pt}{9pt}\selectfont]
    [query]
    \{\{$query\}\}

    Create 5 messages that you think would come up as search results
    if I were to search for messages that match the query. 
    The messages should be very natural, colloquial, and provided
    in bullet type.
    Answers should be of varying lengths, including short sentences of 
    two to three words and longer sentences using up to 10 words.

    your answers:
\end{Verbatim}

\subsection{GPT evaluation prompts}
\label{gpt_eval}
\begin{Verbatim}[commandchars=\\\{\}, fontsize=\fontsize{8pt}{9pt}\selectfont]
\small{hyperparameter} : \{
  "model": "gpt-3.5-turbo-0125",
  "temperature": 0.0,
  "max_tokens": 128
\}
\end{Verbatim}

\textbf{system prompt}
\begin{Verbatim}[commandchars=\\\{\}, fontsize=\fontsize{6pt}{9pt}\selectfont]
    You are a competent bot that can look at a [conversation] 
    and determine whether the user at the 
    end of the conversation is satisfied or not.
    Please answer "satisfied", "dissatisfied", or "neutral". 
    Don't answer anything else.
    
    For each of the following criteria
      satisfied : The assistant's answer meets the user's needs 
      and the user feels satisfied.
      dissatisfied : The user's needs are not yet met and they 
      feel dissatisfied.
      neutral : neither of the above two cases, or simply 
      informational or greeting.

    [conversation]
    \{\{$conversation\}\}

    answer:
\end{Verbatim}

For few-shot, we added one sample each of SAT, NEU, and DSAT from the train dataset to the context.

\section{Interpretability}
\label{apdx:interpretability_b}
\subsection{Additional examples demonstrating feature $F$ interpretability}
\label{additional_examples}
Table \ref{tab:user_utterances} illustrates how the feature values of individual user utterances can provide insights into the reasons behind the predicted satisfaction level, further highlighting the explanatory power of the PRAISE approach.

\captionsetup[table]{skip=6pt}
\begin{table*}[t]

\centering
\begin{tabularx}{\textwidth}{|X|}
\hline
\textbf{Conversation} \\
\textbf{User}: I am trying to find a scary film to watch...I like a lot of styles but today I fancy a scary one \\ \textbf{Assistant}: Have you seen "It (2017)"? \\ \textbf{User}: That's a bit too scary for me. \\ \textbf{Assistant}: How about "Dracula Untold (2014)" ? I hear it's pretty good \\ \textbf{User}: oooh, i have not heard of "Dracula Untold (2014)" ...that sounds good...I like vampire movies. \\
\hline 
\textbf{Label} : SAT \\
\hline 
\textbf{Reason (Score)} \\
User expresses joy over movie discovery. (1.82) \\ User shows interest in movie plot. (1.77) \\ User shows interest in the assistant’s opinion. (1.02) \\
\hline \hline

\textbf{Conversation} \\
\textbf{User} : Do they offer daily housekeeping? \\ \textbf{Assistant} : Daily housekeeping is not available at this hotel. Would you like me to book a room for you? \\ \textbf{User} : Yes that will work, please book. \\
\hline 
\textbf{Label} : SAT \\
\hline 
\textbf{Reason (Score)} \\
User appreciates assistant's recommendation (1.25) \\ User expresses eagerness to use service (1.17) \\ User thanks assistant for patience (1.14) \\
\hline \hline
\textbf{Conversation} \\
\textbf{User} : Thank you. Can you please confirm for me that the guesthouse you booked for me is moderately priced? \\ \textbf{Assistant} : It is actually cheaply priced. Is that going to be okay? \\ \textbf{User} : No, sorry. I am looking for something moderately priced. \\
\hline 
\textbf{Label} : DSAT \\
\hline 
\textbf{Reason (Score)} \\
User expresses confusion about service details (1.04) \\ User shows frustration over lack of information (0.84) \\ User inquires about alternatives for service (0.82) \\
\hline \hline
\textbf{Conversation} \\
\textbf{User} : That suits me well. Can you tell me the address of the venue? \\ \textbf{Assistant} : Your tickets have been bought. Enjoy your time at the event! The address of the venue is 24 Willie Mays Plaza. \\ \textbf{User} : Can you tell me where will the event happen, and at what time it starts? \\
\hline 
\textbf{Label} : NEU \\
\hline 
\textbf{Reason (Score)} \\
User requests additional information (1.30) \\ User expresses excitement about the trip details (1.11) \\ User confirms booking details eagerly (1.06) \\
\hline
\end{tabularx}
\caption{Examples of user utterances}
\label{tab:user_utterances}
\end{table*}

\subsection{Additional box-plot analysis of relevance scores for strategies}
\label{additional_boxplot}

\begin{figure*}[t]
\begin{adjustbox}{width=1.00\textwidth,center}
\begin{minipage}[t]{0.62\textwidth}
    \vspace{0pt} 
    \centering
    \includegraphics[width=\textwidth]{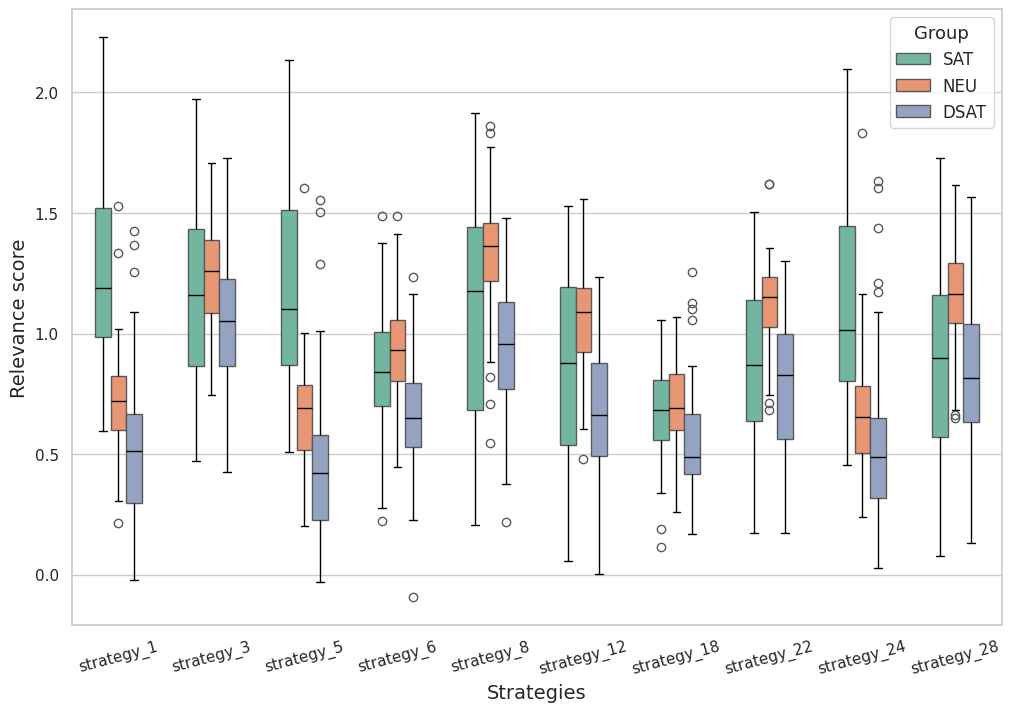}
\end{minipage}%
\hfill
\begin{minipage}[t]{0.38\textwidth}
    \vspace{10pt} 
    \tiny 
    \begin{itemize}[leftmargin=*]
        \item strategy\_1 : User shows gratitude for discovery.
        \item strategy\_3 : User expresses surprise at a movie fact.
        \item strategy\_5 : User expresses satisfaction with conversation.
        \item strategy\_6 : User imagines a movie from the perspective of a minor character.
        \item strategy\_8 : User mentions rewatching a recommended movie.
        \item strategy\_12 : User correlates movie theme with a favorite song.
        \item strategy\_18 : User shows interest in the assistant's opinion.
        \item strategy\_22 : User asks for more details about a movie.
        \item strategy\_24 : User shows gratitude for detailed explanation.
        \item strategy\_28 : User expresses interest in movie plot.
    \end{itemize}
\end{minipage}
\end{adjustbox}
\caption{Box-plot of relevance scores for strategies in the ReDial dataset}
\label{fig:box_redial}
\end{figure*}

\begin{figure*}[t]
\begin{adjustbox}{width=1.00\textwidth,center}
\begin{minipage}[t]{0.62\textwidth}
    \vspace{0pt} 
    \centering
    \includegraphics[width=\textwidth]{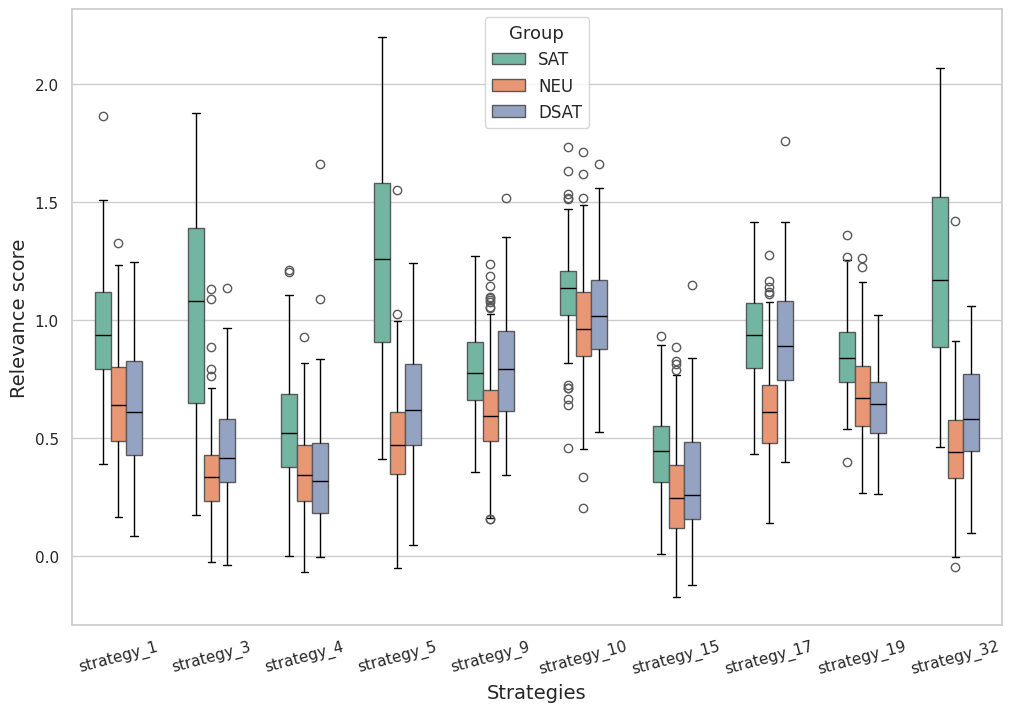}
\end{minipage}%
\hfill
\begin{minipage}[t]{0.38\textwidth}
    \vspace{10pt} 
    \tiny 
    \begin{itemize}[leftmargin=*]
        \item strategy\_1 : User expresses eagerness to use service.
        \item strategy\_3 : User expresses relief after receiving information.
        \item strategy\_4 : User indicates intention to follow advice.
        \item strategy\_5 : User is relieved to find [desired information].
        \item strategy\_9 : User asks for clarification on [specific detail].
        \item strategy\_10 : User expresses confusion about [service details].
        \item strategy\_15 : User acknowledges assistant's thoroughness in response.
        \item strategy\_17 : User requests further clarification.
        \item strategy\_19 : User shows interest in future services.
        \item strategy\_32 : User expresses relief at finding [information].
    \end{itemize}
\end{minipage}
\end{adjustbox}
\caption{Box-plot of relevance scores for strategies in the MWOZ dataset}
\label{fig:box_mwoz}
\end{figure*}

In Figure \ref{fig:box_redial}, strategies that directly express satisfaction or gratitude, such as strategy\_1 and strategy\_5, exhibit a clear distinction in the SAT label. On the other hand, strategy\_22, which involves asking for more details, shows higher scores in the NEU label compared to both SAT and DSAT.

Figure \ref{fig:box_mwoz} demonstrates that, similar to other datasets, strategies with direct positive expressions yield high scores in the SAT label. Notably, strategy\_17, which represents cases where users request further clarification, serves as a clear criterion for distinguishing the NEU label.

\section{Impact of initial strategy selection}
\label{multiple_run}
Table \ref{tab:multiple_run} shows performance comparison across 5
sets of 5 randomly generated strategies for satisfaction classification, each trained 20 times over 50 steps.

\begin{table}[H]
\centering
\scriptsize

\begin{tabular}{|c|c|c|c|c|c|c|}
\hline
\textbf{step} & \textbf{set\_1} & \textbf{set\_2} & \textbf{set\_3} & \textbf{set\_4} & \textbf{set\_5} & \textbf{p\_value} \\
\hline
1 & 0.5239 & 0.5349 & 0.5316 & 0.5244 & 0.5272 & 0.069874 \\
2 & 0.5397 & 0.5508 & 0.5509 & 0.5452 & 0.5415 & 0.023499 \\
3 & 0.5540 & 0.5618 & 0.5618 & 0.5609 & 0.5510 & 0.002639 \\
4 & 0.5631 & 0.5654 & 0.5675 & 0.5661 & 0.5589 & 0.109610 \\
5 & 0.5660 & 0.5667 & 0.5690 & 0.5697 & 0.5638 & 0.429035 \\
\hline
\multicolumn{7}{|c|}{...} \\
\hline
46 & 0.5900 & 0.5903 & 0.5941 & 0.5928 & 0.5920 & 0.480261 \\
47 & 0.5914 & 0.5911 & 0.5939 & 0.5940 & 0.5928 & 0.679273 \\
48 & 0.5911 & 0.5915 & 0.5955 & 0.5941 & 0.5940 & 0.378739 \\
49 & 0.5916 & 0.5908 & 0.5952 & 0.5939 & 0.5930 & 0.447692 \\
50 & 0.5917 & 0.5907 & 0.5953 & 0.5948 & 0.5928 & 0.370706 \\
\hline
\end{tabular}
\caption{Performance comparison across different sets and steps}
\label{tab:multiple_run}
\end{table}

\section{Different models for score analyzer}
The core requirements for the Score Analyzer in the PRAISE framework are interpretability and a simple model capable of rapid training. We selected Random Forest and Logistic Regression as suitable models for these requirements and conducted experiments. The experimental results using Random Forest showed a 5$\sim$10$\%$ decrease in performance compared to our final PRAISE implementation. Based on these results, we ultimately selected Logistic Regression as it demonstrated superior performance while maintaining comparable interpretability to Random Forest.

\section{License}
The licenses for the datasets and baselines are as follows: ReDial (CC-BY-4.0), MWOZ (MIT License), SGD (CC-BY-SA-4.0), USDA (CC-BY-4.0), ASAP (MIT License).

\end{document}